# Deep Learning for Real Time Satellite Pose Estimation on Low Power Edge TPU

Alessandro Lotti, Dario Modenini, Paolo Tortora, Massimiliano Saponara, and Maria A. Perino

*Abstract*— Pose estimation of an uncooperative space resident object is a key asset towards autonomy in close proximity operations. In this context monocular cameras are a valuable solution because of their low system requirements. However, the associated image processing algorithms are either too computationally expensive for real time on-board implementation, or not enough accurate. In this paper we propose a pose estimation software exploiting neural network architectures which can be scaled to different accuracy-latency trade-offs. We designed our pipeline to be compatible with Edge Tensor Processing Units to show how low power machine learning accelerators could enable Artificial Intelligence exploitation in space. The neural networks were tested both on the benchmark Spacecraft Pose Estimation Dataset, and on the purposely developed Cosmo Photorealistic Dataset, which depicts a COSMO-SkyMed satellite in a variety of random poses and steerable solar panels orientations. The lightest version of our architecture achieves state-of-the-art accuracy on both datasets but at a fraction of networks complexity, running at 7.7 frames per second on a Coral Dev Board Mini consuming just 2.2W.

## I. Introduction

Vision based navigation is a key technology for next generation of On-Orbit Servicing (OOS) and Active Debris Removal (ADR) missions. In these scenarios, guidance and control laws shall be fed with the relative chaser-to-target pose (i.e. position and attitude) which might be conveniently estimated from monocular images as these sensors are simple, light, and consume little power. Traditionally, Image Processing (IP) algorithms are divided in i) hand-crafted features [1,2] and ii) Deep Learning (DL) based [3-14]. However, the former are affected by low robustness against typical space imagery characteristics such as low Signal-to-Noise-Ratio (SNR), severe, and rapidly varying illumination conditions, and backgrounds. Neural Networks (NNs) could overcome such weaknesses through proper training but often result in a high computational burden, hardly compatible with typical onboard processing power.

In recent years deep learning has been shown to aid spacecraft monocular pose estimation at different levels. Sharma *et al* [3] addressed the problem as a classification task by discretizing the pose space. Later, Sharma and D'Amico [4], proposed a solution based on joint classification and regression. In this last work, the authors also presented the Spacecraft PosE Estimation Dataset (SPEED), providing synthetic and real high resolution (1920x1200 px) images of the Tango spacecraft. SPEED has been adopted as a benchmark in the first Satellite Pose Estimation Challenge (SPEC) [15] co-hosted by the European Space Agency (ESA) and Stanford's Space Rendezvous Laboratory (SLAB). Three out the four top-scoring works adopted a three stage approach leveraging Convolutional Neural Networks (CNNs) to detect the target on the image first, and to regress the locations of some predefined keypoints later which are then fed into off-the-shelf Perspective-n-Point (PnP) solvers [5,6] for pose estimation. On the other hand, the third classified [7] proposed an end-to-end CNN based pipeline which however required a huge model to achieve competitive accuracy. Indeed, submissions were ranked solely on a pose regression error, regardless of their computational burden. Only the SLAB baseline [6] addressed this issue by adopting light networks which however led to a pose estimation error significantly higher than the best one.

Later works addressed the pose estimation accuracy-latency trade-off. To this aim, Hu *et al.* [8] proposed a single stage method leveraging a 3D loss to make pose estimation less sensitive to scale variations. Wang *et al.* [9] exploited Transformers for direct keypoints coordinates retrieval upon a CNN based detection step, while Piazza *et al.* [10] adopted a light model for target detection. Carcagnì *et al.* [11] revisited the method in [5] by replacing the landmarks regression network backbone with a lighter variant. Similarly, Posso *et al.* [12] revisited the end-to-end approach proposed in [7] in a lite manner, reaching however an accuracy that is far from the top submissions at SPEC.

These works have been tested on high end desktop GPUs only and they are not optimized for low power embedded devices which typically perform well on limited sets of operations as a consequence of hardware specialization. Black *et al.* [13] contributed remarkably in this respect, by developing a light pipeline that achieves real time inference on an Intel Joule 570x board consuming 3.7 W.

In this context, our work focuses on further improving the pose estimation accuracy-latency trade-off on low power devices from both the software and hardware points of view.

First, we propose a pose estimation pipeline based on NNs optimized for embedded devices, with an architecture that can be scaled to the available computational power. We investigate optimizations through TensorFlow Lite (TFLite) conversion and quantization. The former is a Machine Learning (ML) library for on-device inference while the latter consists of

A. Lotti is with the Department of Industrial Engineering, Alma Mater Studiorum Università di Bologna, Forlì, 47121, Italy (e-mail: alessandro.lotti4@unibo.it).

D. Modenini, and P. Tortora are with the Department of Industrial Engineering and with the Interdepartmental Centre for Industrial Research, Alma Mater Studiorum Università di Bologna, Forlì, 47121, Italy (e-mail: {dario.modenini, paolo.tortora}@unibo.it).

M. Saponara and M. A. Perino are with Thales Alenia Space Italia, Turin, 10146, Italy (e-mail: {massimiliano.saponara, mariaantonietta.perino}@thalesaleniaspace.com).

This work was partially supported by Thales Alenia Space Italia, in the framework of the project "A testbed for deep learning in support of docking-grasping operations".

converting NNs' weights to integers, yielding to a significant runtime advantage at deployment.

Second, we investigate high performing ML coprocessors for low power integer-only inference, namely Edge Tensor Processing Units (TPUs), which are gaining the attention of the space community [16]. We thus test our models on a Coral Dev Board Mini equipped with both a TPU and a 1.5 GHz quadcore CPU. We evaluate our algorithms on both SPEED [4] and on a new dataset developed as part of this work, named COSMO Photorealistic Dataset (CPD), depicting a COSMO SkyMed satellite. Our results show how model optimization and edge processors can enable sub-degree and centimeter-level real time pose estimation compatible with typically available onboard power levels.

## II. Cosmo Photorealistic Dataset

Moving components, such as antennas and solar arrays, are common for large spacecrafts, i.e. the ones that would benefit the most from OOS. The need to validate pose estimation pipelines even in this scenario, drove the design of a new dataset, named COSMO Photorealistic Dataset (CPD), depicting a satellite from the SkyMed Earth observation constellation in heterogeneous combinations of poses, solar arrays configurations, lighting conditions, and backgrounds. To this end, the 3D computer graphics software Blender was selected because of its native support to Physically Based Rendering (PBR). Most assumptions adopted for setting the spacecraft pose distribution and image post-processing follow that of SPEED [4] for ease of benchmarking, as detailed in the following.

### A. Blender Scene

The Blender scene consists of three concentric highly polygonal spheres representing Earth, clouds, and atmosphere plus a CAD model of the COSMO spacecraft. Similarly to [7], the Earth was textured with a high resolution map further augmented with an ocean mask[1] and topography data from the NASA's Blue Marble collection[2] which also provided clouds texture. A third-party shader[3] was applied to increase the realism of the clouds providing at the same time transparency, diffusion, and reflection. Atmospheric scattering was emulated exploiting Blender's volumetric rendering in combination with a second shader, from the same package, which implements an exponential density model. A Blender's sun lamp emulates the Sun by providing collimated light at a blackbody temperature of 5778 K.

The main exterior features of COSMO spacecraft are the Multi-Layer Insulation (MLI) and the solar panels. A faithful representation of the MLI material was obtained thanks to a crumpled normal texture[4] mapped to the spacecraft body for emulating the typical random reflections. The solar arrays have been equipped with a solar cell texture providing base color and displacement, while surface reflection has been obtained through a Blender's glossy shader.

### B. Pose Distribution

The distance is randomly selected from a standard normal distribution $N(\mu = 36m, \sigma = 10m)$, rejecting all the values above 70m and below 36m. The x-y offsets on the image plane are uncorrelated random values selected from a multivariate normal distribution, constrained to guarantee that the satellite almost always lies entirely in the image frame. The attitude is randomly sampled from a uniform distribution of rotations in the SO(3) space.

Pose distribution and camera-satellite alignment are governed by the Starfish library [14]. Domain variation is provided through Blender Python API, by rotating Earth and clouds beneath the satellite before each rendering. Geometrical constraints on the Sun-satellite-Earth angle and Sun-satellite-camera angle are prescribed to avoid dark images due to a non-illuminated camera view. Besides that, lighting direction is randomized across the dataset to provide a comprehensive range of illumination conditions. Sun-tracking rotation of solar panels is added through a Blender "locked track" constraint which allow them to rotate about their longitudinal axis for tracking the Sun direction.

### C. Render Setup and Post Processing

A total of 15000 images have been rendered with the PBR Cycles engine through a pinhole camera model. The resolution has been set to 1920x1200 px. Post-processing steps include a glare node, meant to replicate the bloom effect, grayscale conversion, and addition of Gaussian noise and Gaussian blurring to emulate shot noise and depth of field. Fig. 1 depicts a close-up render of the satellite and two samples from the dataset, prior to grayscale conversion and noise addition.

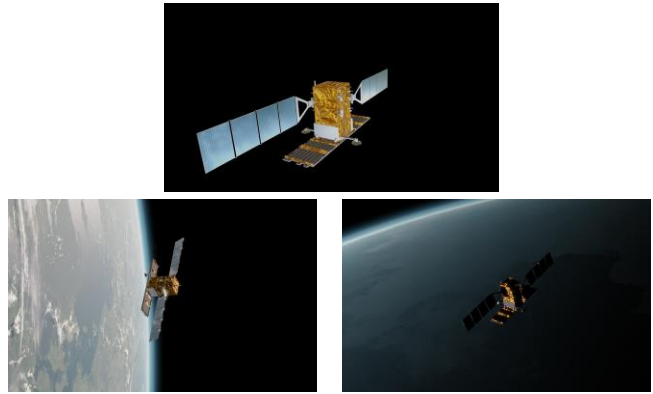

**Fig. 1 Close-up preview of the COSMO SkyMed satellite (top), sample images from CPD prior to postprocessing (bottom)**

## III. Methods

Our software pipeline is based on three stages, namely A) spacecraft detection, B) landmarks regression, and C) pose estimation. For this work, we hold the assumption that the target is known by the chaser, i.e. a wireframe model is available. Since this information was not included in the SPEED [4] dataset, we reconstructed a 3D model through multiview triangulation. We then retrieved training and testing labels (i.e. bounding boxes and landmarks coordinates) for both datasets by projecting the 3D satellites' keypoints, highlighted in Fig. 2, onto the image plane exploiting the known target pose.

---

[1] Tom Patterson, www.shadedrelief.com
[2] NASA, Visible Earth, visibleearth.nasa.gov
[3] F. Lasse, Physically Correct Atmosphere Shader, gumroad.com/l/JINTt
[4] NASA, 3D Resources, nasa3d.arc.nasa.gov/detail/Sentinel-6

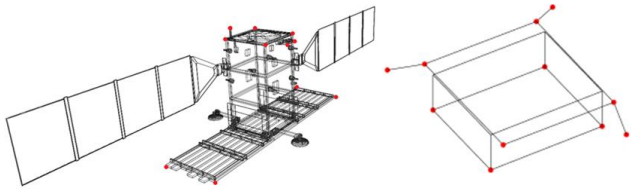

**Fig. 2 Wireframe models: COSMO (left), Tango (right)**

### A. Spacecraft Detection

Direct processing of high resolution images would prevent real-time inference on low power embedded devices, due to the need of large NNs and a high memory footprint. The purpose of the Detection Network (DN) is therefore to identify a Region of Interest (ROI), by detecting the satellite on the image.

To this end, we employed a MobileDet Edge TPU optimized model [17] from the TensorFlow (TF) Object Detection API[5] with an input shape of 320x320 px and about 3.3 millions of parameters. The predicted bounding box is made squared, to avoid distortion, and enlarged by a factor 1.15 to increase the chance that the satellite lies within its margins. The ROI accuracy is defined as the percentage of times the ground truth bounding box is contained within the regressed one. In addition, the matching between the true and estimated ROI is assessed in terms of Intersection Over Union (IoU). In case the ROI is smaller than the second NN input size, the bounding box is further expanded, otherwise the cropped image is resized to fulfil that requirement.

### B. Landmarks Regression

The resulting ROI is fed into a second CNN which is in charge of detecting a set of predefined keypoints. We propose a Regression Network (RN) architecture which can be scaled to different accuracy-latency trade-offs. It consists of a fully convolutional model built on top of EfficientNet-Lite backbones[6,7], which are obtained by removing operations not well supported by mobile accelerators from the original EfficientNets [18]. The regression head consists of a sequence of four convolutions (Table 1) which gradually reduce the feature map channels and dimensions through learnt operations rather than pooling. The network returns a vector containing normalized coordinates of the landmarks in a predefined order.

Model scaling is inherent to the adoption of EfficientNet-Lite backbones which are developed to this purpose, each model featuring a prescribed combination of network's width, depth, and input resolution[8]. In this work, we tested EfficientNet-Lite versions 0 to 4, for a total of 5 variants.

**Table 1 Regression head structure**

| # | Type | Activation | Kernel size | Padding | #filters |
|---|------|------------|-------------|---------|----------|
| 1 | Standard | Relu | 1 | Valid | 128 |
| 2 | Separable | Relu | 3 | Same | 128 |
| 3 | Standard | - | 1 | Valid | kpts x2 |
| 4 | Standard | - | #3 resolution | Valid | kpts x2 |

Target position and orientation are estimated through an EPnP solver [19] with RANSAC exploiting the known 2D-3D correspondences. The resulting pose is eventually refined with a Levenberg–Marquardt optimization step[9].

## IV. Experiments and Results

Our models are first compared with state-of-the-art methods using the floating point, non-optimized NNs configuration. We then illustrate the benefits obtained through TensorFlow Lite[10] conversion first and quantization later. These modifications enable the deployment of the models on a low power machine learning accelerator, namely a TPU, which can further reduce the inference time.

### A. Training Setup

For SPEED dataset, since the ground truth poses for the test sets have not been released at the time of this writing, only synthetic images from the SPEC training set were used. Both SPEED and CPD have been divided into train and test clusters through uniform random sampling (Table 2).

The DN is trained for 50000 steps with momentum optimizer starting from COCO [20] pretrained weights, applying random crops and horizontal flips to avoid overfitting. RNs have been trained with Adam optimizer [21] and mean absolute error loss for 550 epochs on SPEED and 450 epochs on CPD. EfficientNet-Lite backbones were initialized with Imagenet [22] checkpoints. Applied data augmentations include random image rotations, bounding box enlargements and shifts, random brightness, and contrast adjustments.

The learning rate was gradually reduced according to a cosine decay law after a warmup phase, lasting 2000 steps for the DN and 10 epochs for the RNs, where it grows linearly from 0.15 to 0.45 and from 1e-4 to 3e-3 respectively. The batch size was set to 256 for both DN and RNs.

The trained networks were first deployed on the CPU of a Coral Dev Board Mini for performance assessment. Later, all RNs have been re-trained exploiting Quantization Aware Training[11] (QAT) which emulates quantization along with NNs parameters tuning, to reduce accuracy loss at conversion. The only exception is the DN which is trained only once for each of the two datasets directly applying QAT, to reduce development time.

The results provided in next paragraphs refer to the SPEED [4] dataset, unless otherwise stated.

**Table 2 Datasets partitioning**

| Dataset | Train images | Test images |
|---------|--------------|-------------|
| SPEED | 9728 | 2272 |
| CPD | 12032 | 2968 |

### B. Comparison with State of the Art

To assess the accuracy of our methods we adopt the same metric of SPEC [15]. This is based on the sum of a normalized position ($\bar{e}_t$) and rotation ($e_q$) errors averaged over all the $N$ test images:

---

[5]github.com/tensorflow/models/blob/master/research/object_detection/g3doc/tf1_detection_zoo.md
[6]blog.tensorflow.org/2020/03/higher-accuracy-on-vision-models-with-efficientnet-lite.html
[7]github.com/sebastian-sz/efficientnet-lite-keras
[8]github.com/tensorflow/tpu/blob/master/models/official/efficientnet/lite/efficientnet_lite_builder.py#L38-L45
[9]docs.opencv.org/3.4.12/d9/d0c/group__calib3d.html
[10]tensorflow.org/lite?hl=en
[11]tensorflow.org/model_optimization/guide/quantization/training?hl=en

$$e_{t_i} = |t_{GT_i} - t_{EST_i}|_2$$
$$\bar{e}_t = \frac{1}{N}\sum_{i=1}^{N} \frac{e_{t_i}}{|t_{GT_i}|_2}$$
$$e_q = \frac{1}{N}\sum_{i=1}^{N} 2\arccos(|\langle q_{GT_i}, q_{EST_i}\rangle|)$$
$$E = \bar{e}_t + e_q$$

Where $t_{GT}$, $q_{GT}$, and $t_{EST}$, $q_{EST}$ denote the ground truth and estimated position and attitude quaternion respectively. Table 3 displays the error metrics attained by all of our pipeline variants along with those of the best ranked submissions at SPEC and the embedded method proposed in [13]. Even though a direct comparison with would require to train and test the NNs on the same images subsets, our results clearly indicate how state-of-the art accuracy can be achieved with remarkably less NNs parameters.

**Table 3 Comparison with state of the art [6-8,13,15]**

| Model | E | $e_q$ [°] | $e_t$ [m] | #Parameters |
|---|---|---|---|---|
| UniAdelaide [5] | 0.0094 | 0.41 ± 1.50 | 0.032 ± 0.095 | 176.2 M |
| Our_lite4 | 0.0119 | 0.52 ± 0.52 | 0.034 ± 0.069 | 15.4 M |
| Our_lite3 | 0.0124 | 0.55 ± 0.53 | 0.033 ± 0.062 | 10.5 M |
| Our_lite2 | 0.0131 | 0.58 ± 0.57 | 0.036 ± 0.075 | 8.4 M |
| Our_lite1 | 0.0134 | 0.59 ± 0.57 | 0.037 ± 0.068 | 7.7 M |
| Our_lite0 | 0.0149 | 0.65 ± 0.58 | 0.040 ± 0.073 | 6.9 M |
| EPFL_cvlab[12] | 0.0215 | 0.91 ± 1.29 | 0.073 ± 0.587 | 89.2 M |
| Black, et al. [13] | 0.0409 | - | - | 6.9 M |
| pedro_fairspace [7] | 0.0571 | 2.49 ± 3.02 | 0.145 ± 0.239 | ≈ 500 M |
| SLAB_baseline [6] | 0.0626 | 2.62 ± 2.90 | 0.209 ± 1.133 | 11.2 M |

### C. Conversion to TensorFlow Lite

As TF is not optimized for on-device inference, we converted all NNs to TFLite. In this work, we exploited the 2.7.0 version of tflite-runtime. Performance comparisons are reported in Table 4 and Table 5, highlighting latency reduction and persistence of accuracy. Note that the frames per second (fps) data do not include image loading time from memory.

**Table 4 DN performances**

| DN | IoU mean | IoU median | ROI accuracy |
|---|---|---|---|
| TF | 0.9426 | 0.9600 | 96.88 % |
| TFLite | 0.9408 | 0.9575 | 97.40 % |

**Table 5 RN variants details and TF vs TFLite pipelines performances comparison**

| RNs | Input size [px] | Pipeline fps | | Pose error E | |
|---|---|---|---|---|---|
| | | TF | TFLite | TF | TFLite |
| Lite0 | 224x224 | 0.69 | 0.76 | 0.0149 | 0.0150 |
| Lite1 | 240x240 | 0.61 | 0.68 | 0.0134 | 0.0136 |
| Lite2 | 260x260 | 0.56 | 0.62 | 0.0131 | 0.0131 |
| Lite3 | 280x280 | 0.46 | 0.51 | 0.0124 | 0.0126 |
| Lite4 | 300x300 | 0.33 | 0.39 | 0.0119 | 0.0118 |

### D. Quantization: CPU vs TPU Inference

Quantization, which is needed for deploying the networks on the TPU, allows compressing the file sizes up to the 75% thereby significantly reducing latency. Accuracy drop is inevitable; however, to minimize it, RNs have been retrained applying QAT. All networks have been fully quantized except for input and output tensors. A comparison between CPU and TPU performances is provided in Table 6 and Table 7, highlighting the superior fps attainable by the latter, which is up to the 253% higher than with the CPU. In addition, the TPU allows reducing power consumption at inference: our tests revealed that the average absorption drops from the 3 W required by CPU to 2.2 W for TPU.

**Table 6 CPU vs TPU quantized NNs runtime**

| RNs | DN Runtime [ms] | | RN Runtime [ms] | |
|---|---|---|---|---|
| | CPU | TPU | CPU | TPU |
| Lite0 | 303.54 | 69.82 | 136.19 | 36.77 |
| Lite1 | 300.58 | 97.92 | 194.29 | 63.26 |
| Lite2 | 296.76 | 128.26 | 261.83 | 92.45 |
| Lite3 | 319.60 | 139.39 | 425.67 | 177.49 |
| Lite4 | 304.23 | 144.15 | 658.96 | 414.77 |

**Table 7 CPU vs TPU quantized pipelines performances**

| RNs | Pose error, E | Board Temperature [°C] | | Pipeline speed [fps] | | |
|---|---|---|---|---|---|---|
| | | CPU | TPU | CPU | TPU | Gain [%] |
| Lite0 | 0.0179 | 83.82 | 51.86 | 2.17 | 7.66 | +253% |
| Lite1 | 0.0156 | 84.63 | 51.66 | 1.94 | 5.38 | +177% |
| Lite2 | 0.0151 | 84.80 | 51.43 | 1.72 | 4.06 | +136% |
| Lite3 | 0.0152 | 84.79 | 48.95 | 1.30 | 2.92 | +125% |
| Lite4 | 0.0147 | 85.39 | 46.71 | 1.01 | 1.70 | +68% |

### E. Performances on CPD

Finally, we tested our TPU pipelines on CPD. The good performance of the DN (Table 8) demonstrates its robustness against variable solar panels orientation. The mean pose error $E$ is similar to that on SPEED, even though few outliers are present, as highlighted in Table 9. The worst results are obtained on images where the solar panels occlude a large portion of the SAR antenna, images taken in near eclipse conditions, and those characterized by extreme blooming.

**Table 8 DN performances on CPD**

| Quantized DN | IoU mean | IoU median | ROI accuracy |
|---|---|---|---|
| | 0.9417 | 0.9537 | 99.2 % |

**Table 9 TPU pipelines performances on CPD**

| Quantized RNs | E | $e_q$ [°] | $e_t$ [m] |
|---|---|---|---|
| Lite0 | 0.0170 | 0.741 ± 0.804 | 0.180 ± 0.355 |
| Lite1 | 0.0153 | 0.667 ± 0.638 | 0.161 ± 0.230 |
| Lite2 | 0.0159 | 0.654 ± 1.648 | 0.202 ± 2.644 |
| Lite3 | 0.0141 | 0.613 ± 0.589 | 0.152 ± 0.273 |
| Lite4 | 0.0140 | 0.580 ± 0.660 | 0.173 ± 1.598 |

## V. Conclusion

We introduced a new photorealistic satellite dataset including steerable solar panels and we discussed neural models enabling real-time spacecraft pose estimation from monocular images on low power embedded hardware.

Our pipelines perform on par with the state of the art while using extremely lite networks. Switching from CPU to TPU allows increasing the fps by a factor of ≈ 2 to 3 (up to 7.7 fps) while reducing the measured power consumption of 25% (from 3W down to 2.2 W).

When evaluated on our CPD, the algorithms exhibit accuracies in line with SPEED, although with higher sensitivity to occlusions, which deserves further investigation.

---

[12] indico.esa.int/event/319/attachments/3561/4754/pose_gerard_segmentation.pdf

Future developments include testing the NNs on real imagery to investigate domain gap.